\documentclass{article}

\usepackage{microtype}
\usepackage{graphicx}
\usepackage{subcaption}
\usepackage{booktabs} 
\usepackage{multirow} 

\usepackage{hyperref}

\usepackage[preprint]{icml2026}

\usepackage{amsmath}
\usepackage{amssymb}
\usepackage{mathtools}
\usepackage{amsthm}
\usepackage{enumitem}
\usepackage{wasysym}
\usepackage{pifont}
\usepackage[dvipsnames]{xcolor}

\usepackage{listings}
\usepackage[capitalize,noabbrev]{cleveref}

\theoremstyle{plain}

\theoremstyle{definition}

\theoremstyle{remark}

\usepackage[disable,textsize=tiny]{todonotes}

\newcommand{\need}{\textcolor{BrickRed}{Need}}
\newcommand{\noneed}{\textcolor{ForestGreen}{No Need}}
\definecolor{red}{RGB}{192,0,0}
\renewcommand{\epsilon}{\varepsilon} 

\icmltitlerunning{DRIFT: Detecting Representational Inconsistencies for Factual Truthfulness}

\begin{document}

\twocolumn[
  \icmltitle{DRIFT: Detecting Representational Inconsistencies for Factual Truthfulness}


  \icmlsetsymbol{equal}{*}
  \icmlsetsymbol{corresp}{$\dagger$}

  \begin{icmlauthorlist}
    \icmlauthor{Rohan Bhatnagar}{equal,umd-cs}
    \icmlauthor{Youran Sun}{equal,umd-math}
    \icmlauthor{Chi Andrew Zhang}{uchicago}
    \icmlauthor{Yixin Wen}{corresp,ufl}
    \icmlauthor{Haizhao Yang}{corresp,umd-cs,umd-math}
  \end{icmlauthorlist}

  \icmlaffiliation{umd-cs}{Department of Computer Science, University of Maryland, College Park, MD, USA}
  \icmlaffiliation{umd-math}{Department of Mathematics, University of Maryland, College Park, MD, USA}
  \icmlaffiliation{uchicago}{Department of Statistics, University of Chicago, Chicago, IL, USA}
  \icmlaffiliation{ufl}{Department of Geography, University of Florida, Gainesville, FL, USA}

  \icmlcorrespondingauthor{Yixin Wen}{yixin.wen@ufl.edu}
  \icmlcorrespondingauthor{Haizhao Yang}{hzyang@umd.edu}

  \icmlkeywords{Hallucination Detection, Large Language Models, Uncertainty Estimation}

  \vskip 0.3in
]

\printAffiliationsAndNotice{\icmlEqualContribution}

\begin{abstract}
LLMs often produce fluent but incorrect answers, yet detecting such hallucinations typically requires multiple sampling passes or post-hoc verification, adding significant latency and cost.
We hypothesize that intermediate layers encode confidence signals that are lost in the final output layer, and propose a lightweight probe to read these signals directly from hidden states.
The probe adds less than 0.1\% computational overhead and can run fully in parallel with generation, enabling hallucination detection before the answer is produced.
Building on this, we develop an LLM router that answers confident queries immediately while delegating uncertain ones to stronger models.
Despite its simplicity, our method achieves SOTA AUROC on 10 out of 12 settings across four QA benchmarks and three LLM families, with gains of up to 13 points over prior methods, and generalizes across dataset shifts without retraining.
\end{abstract}

\section{Introduction}
\label{sec:intro}

Large language models (LLMs) have made remarkable progress recently \cite{qwen3, gemini25} and are being applied in an increasing number of real-world scenarios.
However, hallucination, which refers to situations where a model produces information that appears plausible but is actually false or not supported by facts, undermines users' trust in LLMs and limits their use in critical settings.
Therefore, developing effective methods for hallucination detection is critical.
Ideally, such methods should maintain real-time responsiveness without increasing response latency or significantly increasing generation costs.

Existing methods fall into two categories.
Sampling-based approaches, such as Semantic Entropy \cite{semantic-entropy2}, generate multiple answers and compute semantic uncertainty, but require multiple forward passes, incurring a high computational cost that is incompatible with real-time requirements.
Representation-based approaches, such as HaloScope \cite{haloscope}, train classifiers on final-layer hidden states to predict hallucination risk, but final-layer representations have already been projected toward the token space, potentially discarding internal confidence signals.

We propose a lightweight hallucination-detection method based on intermediate representations that predicts hallucination risk efficiently and accurately.
Unlike prior representation-based methods that assume a fixed low-dimensional truth subspace, our approach detects representational instability before projection to token space.
The additional computation introduced by our method is less than 0.1\% of that required to generate a single token, so our approach adds negligible cost to generation.
Moreover, the detector can be evaluated in parallel with inference, without adding extra latency for the user.
In essence, we aim to build a ``lie detector'' or ``mind reader'' for LLMs.

With an effective hallucination detector, we can build an \textit{LLM router} that adaptively allocates computation.
Given a user query, the default LLM begins generating a response while our detector estimates the risk of hallucination in parallel.
Because the detector can rely solely on the question's intermediate representations, it can run concurrently with generation.
If the predicted risk is low, the system returns the generated response as usual; otherwise, it routes the query to a slower but more reliable pipeline (e.g., a stronger model, reasoning-augmented generation, cross-model verification, or retrieval-augmented generation (RAG)).
In the low-risk case, this design introduces zero additional latency; in the fallback case, the extra delay is less than the time to generate a single token.

\begin{figure*}[ht]
    \centering
    \includegraphics[width=0.75\textwidth]{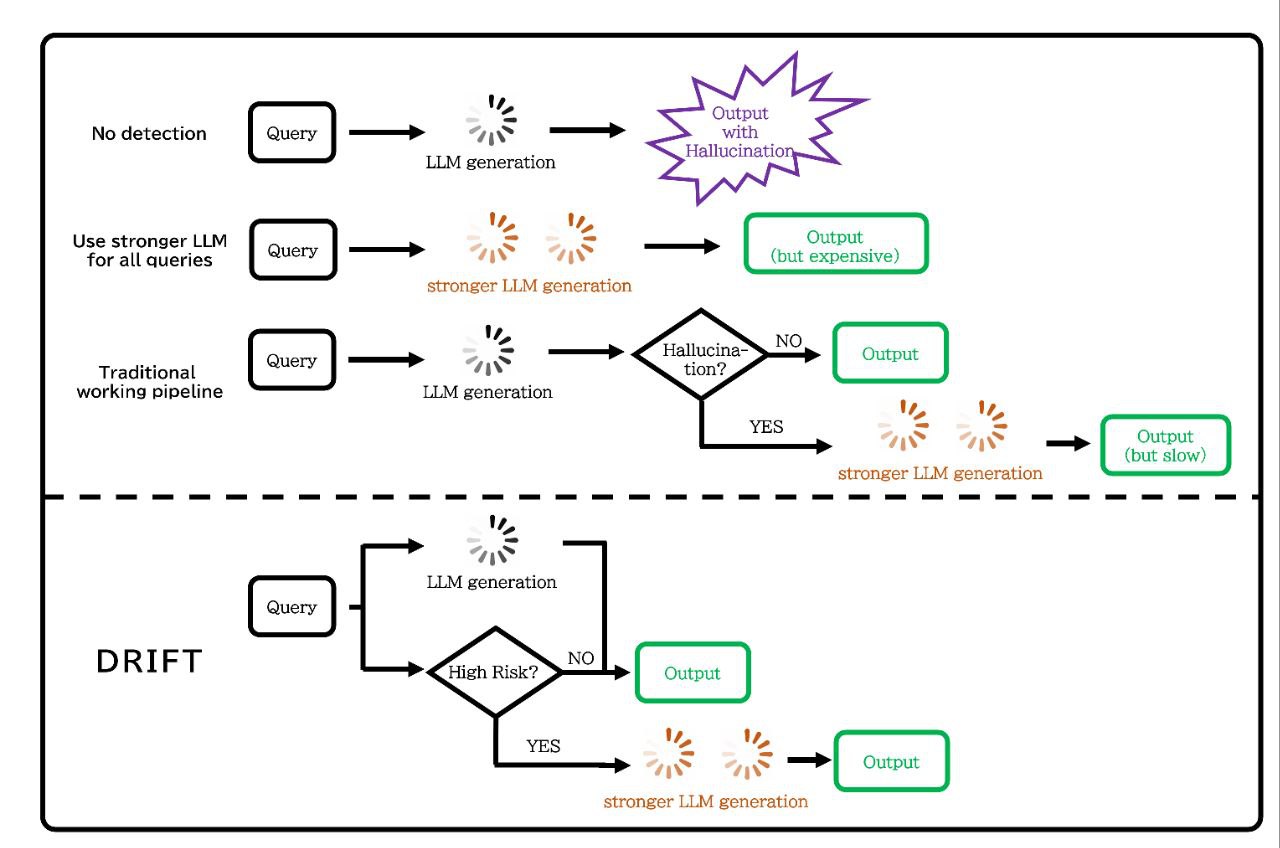}
    \caption{
Comparison of hallucination handling strategies.
Traditional detection pipelines wait for the generation to complete before checking for hallucinations, doubling latency in the fallback case.
Our method evaluates hallucination risk in parallel with generation, enabling zero-latency responses for confident queries and supporting selective prediction and compute allocation while routing uncertain ones to stronger verification pipelines.
}
    \label{fig:pipeline}
\end{figure*}

Our motivation follows naturally from recent evidence that LLMs perform substantial hidden consideration during forward propagation.
Prior work \cite{anthropic2025biology,anthropic2025cot} suggests that intermediate representations encode reasoning, planning, and control signals that guide generation, yet these signals may not be faithfully expressed in the final text.
This motivates directly reading out hallucination-related signals from intermediate representations, rather than relying on generated results.

Based on the above observations, we hypothesize that intermediate representations encode uncertainty signals.
This is analogous to how students taking an exam can sense, for a question, ``I don't know this'' yet would never write that on the answer sheet.
LLMs exhibit a similar pattern because they are trained to provide an answer whenever possible.
This motivates us to build a small detector that captures this internal sense of confusion.
Moreover, this analogy suggests that intermediate layers should be more informative than the final output, just as the internal feeling differs from what appears on the answer sheet.

In our experiments, we find that using representations from intermediate layers is often more effective than using the final-layer representations.
As noted earlier, final-layer representations lose information during projection to token space.
This occurs because the model may contain internal features related to confidence, but since these features are unnecessary for next-token prediction, they are discarded in the last few layers.

Moreover, recent work shows that larger models perform richer latent computation, and their chain-of-thought outputs can be less faithful to internal reasoning~\cite{anthropic2025cot}.
This suggests that the gap between intermediate representations and surface outputs may widen with model scale.
We therefore hypothesize that our method should be more effective when applied to larger models.

Furthermore, our method supports two operating modes with different latency-accuracy trade-offs.
Using only question representations enables prediction before the answer is generated, allowing proactive model switching with minimal latency overhead.
While answer representations yield higher accuracy (as shown in \cref{tab:mainresults}), question-only detection already outperforms prior baselines in most settings, making it a practical choice when low latency is prioritized.

\begin{table*}[!ht]
    \centering
    \caption{Summary of hallucination detection methods.}
    \begin{tabular}{l|cccc}
    \toprule
        Method & Sampling & Features Used & Q/A Features Used \\
    \midrule
        Perplexity & \noneed & Output Logits & Answer \\
        Semantic Entropy \cite{semantic-entropy2} & \need & Output & Answer \\
        Lexical Similarity \cite{Lexical-Similarity} & \need & Output & Answer \\
        SelfcheckGPT \cite{selfcheckGPT} & \need & Output & Answer \\
        EigenScore \cite{EigenScore} & \need & Middle Hidden States & Answer \\
        P(I Know) \cite{languagemodelsmostlyknow} & \noneed & Last Hidden State & Question \\
        True Direction \cite{truthuniversal} & \noneed & Last Hidden State & Answer \\
        HaloScope \cite{haloscope} & \noneed & Middle Hidden States & Answer \\
        HARP \cite{harp} & \noneed & Last Hidden State & Answer \\
        \textbf{Ours} & \noneed & Middle Hidden States & Question/Answer \\
    \bottomrule
    \end{tabular}
    \label{tab:relatedwork}
\end{table*}

Our main contributions are as follows.
\begin{itemize}[itemsep=0pt]
\item We propose a lightweight hallucination detector that reads uncertainty signals from intermediate hidden states. The detector incurs less than 0.1\% computational overhead and can be evaluated in parallel with inference, introducing zero latency.
\item On four QA benchmarks across three LLM families, our method achieves SOTA AUROC on 10 out of 12 settings, with gains of up to 13 points over prior methods, and exhibits strong out-of-distribution generalization across datasets.
\item We develop an LLM router that leverages question-only predictions to route high-risk queries to stronger models before answer generation, improving answer correctness without increasing response latency.
\item Through extensive ablations, we identify that (a) intermediate-layer representations outperform final-layer representations, and (b) our method benefits more from larger models, consistent with our hypothesis.
\end{itemize}

\section{Related Work}

The perplexity of an LLM's answer can itself serve as an indicator for hallucination detection, but as shown in \cite{ren2023outofdistributiondetectionselectivegeneration}, it is unreliable.
\cite{semantic-entropy} and \cite{semantic-entropy2} propose semantic entropy. They require the LLM to generate multiple answers to a given question, then cluster them, and determine the hallucination likelihood based on the entropy of the clusters. Higher semantic entropy indicates a higher probability of hallucination.
Similarly, \cite{Lexical-Similarity}, \cite{selfcheckGPT}, and \cite{EigenScore} detect hallucinations based on consistency over multiple sampled answers.

However, a practical hallucination detector should require much less computation, at least smaller than the cost of generating an answer with the LLM.
\cite{languagemodelsmostlyknow} trained a classifier that uses the last hidden state of the question to predict whether the model knows the answer, i.e., $P(\text{I know})$.
\cite{truthuniversal} used the last hidden state of the answer to learn, via linear regression, two directions, the True direction and the False direction, and performed hallucination detection within this two-dimensional subspace.

Based on the above methods, HARP \cite{harp} and HaloScope \cite{haloscope} employ SVD to project the last or intermediate hidden states corresponding to the answer tokens into a low-dimensional subspace, after which a two-layer MLP is used to regress the hallucination score.
As a follow-up to HaloScope, \cite{steerllmlatentshallucination} attempts to identify hallucinations by modifying the intermediate representations of the LLM and then classifying the last hidden state.

\section{Methodology}
\label{sec:method}

\paragraph{Problem Definition}
We formulate hallucination detection as a \textit{knowledge prediction} task.
Given a user query \( q \), our goal is to estimate whether the LLM can correctly answer it before or during generation.

We focus on closed-book question answering, where the model generates answers solely from its parametric knowledge without access to external sources.
An incorrect answer in this setting indicates that the model has produced content inconsistent with factual ground truth, which aligns with the definition of factual hallucination.
As in prior work \cite{haloscope,semantic-entropy2}, we treat correctness labels as a proxy for hallucination detection.

During the forward process, the LLM produces hidden representations \( h_l \in \mathbb{R}^{N \times D} \) for each layer \( l \in [0, L] \),
where \( N \) is the sequence length, \( D \) is the hidden dimension, and \( L \) is the number of transformer layers.
We use a lightweight auxiliary network \( g_\theta \) that takes the hidden representation \( h \) as input and outputs a scalar score
\[
p = g_\theta(h),
\]
where \( p \in [0,1] \) represents the probability that the model can correctly answer the query \( q \).
A low value of \( p \) indicates that the model is uncertain or likely to hallucinate when generating the response.

\paragraph{Feature Extraction from LLM}

The detector takes hidden representations from a single intermediate layer as input.
We select the layer index as a hyperparameter and find that intermediate layers outperform the final layer (see \cref{fig:layer-ablation}).
Formally, the feature we use can be represented as
\[
h_l = \mathrm{Transformer}_l(q),
\]
where \( h_l \) denotes the hidden representation of the query \( q \) at layer \( l \).

By default, we use only the hidden states corresponding to question tokens, which allows the detector to run before answer generation begins.
When answer tokens are available, including them improves accuracy at the cost of increased latency.

\paragraph{Training Objective}

We construct the training data from question-answering benchmarks that include reference answers.
For each question \( q \) with the standard answer \( a^\ast \), the target LLM generates one answer \( a \).
We then obtain a correctness label by using an external judge to compare \((q,a)\) with \(a^\ast\).
In particular, we use GPT-4o to produce more accurate labels than rule-based evaluation methods.
In principle, multiple answers can be sampled for each question to estimate the probability of correctness.
Still, in practice, we generate only one answer per question due to time and cost limitations.
The resulting label is \( y \in \{0,1\} \), where \( y=1 \) means the generated answer is correct and \( y=0 \) means it is hallucinated or incorrect.
Given the detector output \( p \in [0,1] \), we train the network by minimizing the binary cross-entropy loss between \( p \) and the correctness label \( y \).

\paragraph{Detector Architecture}

\begin{table}[ht]
    \centering
    \caption{Detector architecture configurations. Input dimension varies by base LLM (e.g., 3584 for Qwen-2.5-7B).}
    \begin{tabular}{lccc}
    \toprule
        Architecture & Hidden Dim & Layers & Params \\
    \midrule
        MLP & 128--1024 & 4 & 3M--37M  \\
        Transformer & 256--512 & 4--8 & 4M--30M  \\
    \bottomrule
    \end{tabular}
    \label{tab:architecture}
\end{table}

We experiment with two architectures for the hallucination detector, MLP and Transformer, as summarized in \cref{tab:architecture}.
Both networks take the hidden representations of the question \( h_l \) as input and output a single confidence score \( p \).

For the MLP architecture, the input must have a fixed length.
Therefore the hidden representation \( h_l \in \mathbb{R}^{N \times D} \) is first compressed into a single vector in \( \mathbb{R}^{D} \).
We consider several simple aggregation methods, including mean pooling, max pooling, and using the hidden state of the last token, which often corresponds to the question mark.
We also consider applying principal component analysis (PCA) to the hidden representations and retaining the top $n$ principal components.
The resulting pooled vector or concatenated principal components are then fed into an MLP that predicts the confidence score $p$.

The transformer architecture is more flexible since it naturally processes sequence representations.
In this case, the hidden states \( h_l \) are directly used as input to a lightweight transformer encoder.
This model can capture token-level interactions within the question without requiring input-stage pooling.
At the output stage, we apply attention pooling to aggregate the sequence into a single vector.
Specifically, a small MLP computes a scalar score for each token, and the final representation is a softmax-weighted sum of all token representations.

\section{Experiments}

\begin{table}[ht]
\centering
\caption{Summary of four QA benchmarks used for hallucination detection evaluation. MCQ = multiple-choice question.}
\label{tab:datasets}
\begin{tabular}{lccc}
\toprule
\textbf{Dataset} & \textbf{Size} & \textbf{Type} & \textbf{Topic} \\
\midrule
TriviaQA & 99K & Open QA & Trivia \\
NQ-Open & 92K & Open QA & Wikipedia \\
MMLU-Pro & 12K & MCQ & Professional \\
WebQuestions & 6K & Open QA & Entity \\
\bottomrule
\end{tabular}
\end{table}

\begin{table*}[ht]
    \centering
    \caption{Hallucination detection performance (AUROC, \%) on four QA benchmarks across three model families. Bold indicates best performance per model-dataset pair. * indicates severe label imbalance (see \cref{tab:dataset-statistics}). Our method consistently outperforms baselines in most settings.}
    \begin{tabular}{ll|cccc}
    \toprule
        Model & Method & TriviaQA & NQ-Open & MMLU-Pro & WebQuestions \\
    \midrule
        LLaMA 2 Chat 7B
        & HaloScope(question) & 66.92& 66.97& \textbf{79.56} & 72.62\\
        & HaloScope(answer) & 77.40 & 67.05& 74.83 & 77.03\\
        & Semantic Entropy & 81.23 & 77.57 & 76.76 & 80.12\\
        & Ours (question) & 82.10 & 79.87 &  71.22 & 80.79 \\
        & Ours (answer) & \textbf{88.96} & \textbf{83.76} & 77.85* & \textbf{82.76}\\

    \midrule
        Qwen-2.5-7B-Instruct
        & HaloScope(question) & 85.48& 70.33& 81.08 & 80.43\\
        & HaloScope(answer) & 86.54& 85.84& 77.56& 78.56\\
        & Semantic Entropy & 78.48 & 80.22 & 74.05 & 80.71\\
        & Ours (question) & 87.31 & 83.28 & 82.21 & 84.01\\
        & Ours (answer) & \textbf{93.95} & \textbf{88.43} & \textbf{87.08} & \textbf{87.67} \\

    \midrule
        Gemma-3-4b-it
        & HaloScope(question) &  76.52 & 79.62& 77.73 & 76.70\\
        & HaloScope(answer) & 70.13& 85.84 & \textbf{77.56} & 75.72\\
        & Semantic Entropy & 76.08 & 75.48 & 57.92 & 76.76\\
        & Ours (question) & 83.93 & 82.96 & 74.65 & 79.63 \\
        & Ours (answer) & \textbf{88.80} & \textbf{86.04*} & 75.85* & \textbf{83.11}\\

    \bottomrule
    \end{tabular}
    \label{tab:mainresults}
\end{table*}

\paragraph{Datasets and models}
Our experiments use four open-domain question answering benchmarks, namely TriviaQA \cite{triviaqa}, NQ-Open \cite{nqopen}, MMLU-Pro \cite{mmlupro}, and WebQuestions \cite{webquestions}, summarized in \cref{tab:datasets}.
TriviaQA contains general knowledge trivia questions in an unfiltered, no-context setting.
NQ-Open consists of naturally occurring queries from Google Search paired with Wikipedia-based answers.
MMLU-Pro provides multiple-choice questions spanning 14 professional and academic subjects, often requiring multi-step reasoning.
WebQuestions includes entity-centric questions originally collected from the Google Suggest API.
We evaluate our approach on three model families, namely LLaMA-2 \cite{llama2}, Qwen2.5 \cite{qwen2.5}, and Gemma-3 \cite{gemma3}.
Ablation studies on model scale are conducted using the 7B, 14B, and 32B instruct variants of Qwen2.5.
Detailed statistics on correctness labels for each model-dataset pair are provided in \cref{tab:dataset-statistics}.

\paragraph{Evaluation Metrics}
AUROC (area under the ROC curve) is employed as the primary evaluation metric.
The ROC (receiver operating characteristic) curve plots the true positive rate (TPR) against the false positive rate (FPR) across different classification thresholds.
AUROC summarizes a binary classifier’s ability to separate positives from negatives over all thresholds, ranging from 0 to 1, where higher values indicate stronger discriminative power.

In addition, accuracy and AURAC (area under the rejection–accuracy curve) are also reported.
The RAC plots accuracy against coverage, where samples are ranked by confidence, and coverage denotes the fraction of samples retained.
AURAC assesses whether the model's confidence is reliable, with higher values indicating greater alignment between correctness and confidence.
\cref{fig:rac} provides a concrete example of an RAC curve, which may aid understanding of this metric.

\paragraph{Baselines} We compare our method with the highest performing open baselines, including HaloScope \cite{haloscope} and Semantic Entropy \cite{semantic-entropy2}.

\paragraph{Correctness labeling}
Rule-based correctness labeling methods (e.g., string matching) are unreliable for two reasons.
First, for open-ended questions, semantically equivalent answers may differ in surface form.
Second, when models produce chain-of-thought outputs, intermediate reasoning steps may mention the correct answer, which can lead to mislabeling.
We therefore adopt a more expensive but reliable approach, using GPT-4o to obtain correctness labels.
For each question, we prompt GPT-4o with the original question, the ground-truth answer, and the LLM-generated answer, asking it to return a binary correctness label (see Appendix~\ref{sec:prompt} for the full prompt).
This approach provides a more robust evaluation than rule-based matching.

\subsection{Main Results}
\cref{tab:mainresults} presents the main results.
For each method-model-dataset-feature combination, we use the best-performing layer to ensure a fair comparison.
Our method achieves the best AUROC on 10 out of 12 model-dataset combinations.
Qwen-2.5-7B-Instruct shows the strongest performance, with our approach outperforming all baselines across all four benchmarks, achieving gains of up to 13.0 points (question-only on NQ-Open) and 9.5 points (question-plus-answer on MMLU-Pro).
On LLaMA 2 Chat 7B and Gemma-3-4b-it, we observe similar improvements, surpassing prior methods by 7.7 and 12.3 points on TriviaQA, respectively.
The two exceptions occur on MMLU-Pro for LLaMA 2 and Gemma-3, where HaloScope achieves slightly higher AUROC.
We attribute this to two factors.
First, a format mismatch, as TriviaQA, NQ-Open, and WebQuestions are open-ended QA tasks while MMLU-Pro is a multiple-choice benchmark where the model selects from predefined options rather than generating free-form answers.
Second, label imbalance: for example, LLaMA 2 and Gemma-3 achieve only 15.4\% and 27.1\% accuracy on MMLU-Pro, respectively (see \cref{tab:dataset-statistics}), resulting in limited positive samples for training the detector.
Using answer token representations (Ours answer) consistently outperforms question token representations (Ours question) across all settings, suggesting that answer tokens encode richer information about the model's factual confidence.
Overall, these results demonstrate that intermediate-layer representations provide a strong signal for hallucination detection, competitive with or superior to methods that require sampling multiple outputs.

\begin{figure*}[ht]
    \centering
    \begin{subfigure}[b]{0.45\textwidth}
        \centering
        \includegraphics[width=\textwidth]{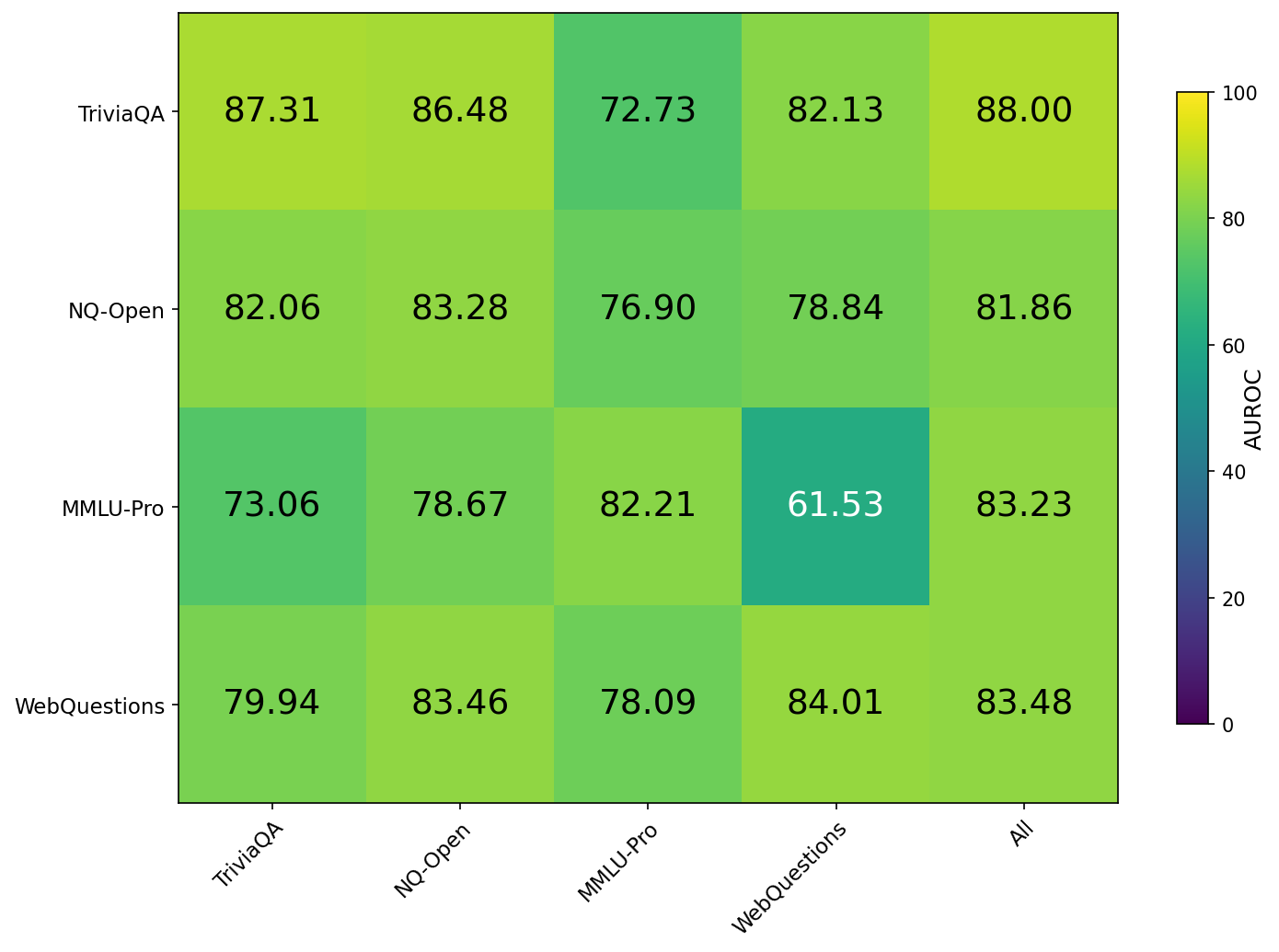}
        \caption{Using question tokens only}
        \label{fig:ood_q}
    \end{subfigure}
    \hfill
    \begin{subfigure}[b]{0.45\textwidth}
        \centering
        \includegraphics[width=\textwidth]{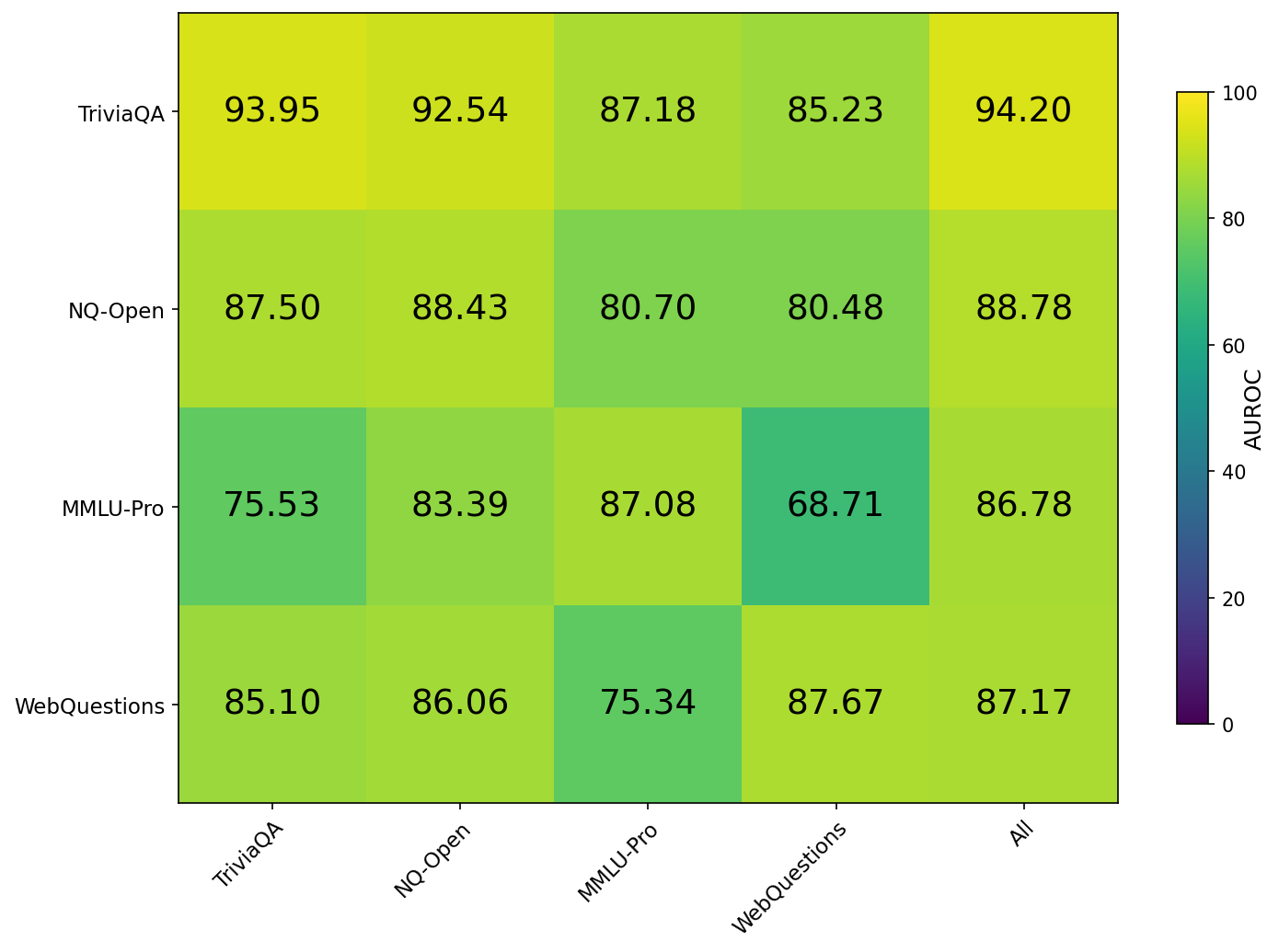}
        \caption{Using question + answer tokens}
        \label{fig:ood_qa}
    \end{subfigure}
    \caption{Out-of-distribution generalization. Heatmap of AUROC when training the detector on one dataset (columns) and evaluating it on another (rows), including training on the union of all datasets (\textit{All}).}
    \label{fig:combined_ood}
\end{figure*}

\subsection{Out-of-Distribution Generalization}

In practice, the distribution of user queries may differ from the training data.
We therefore evaluate whether our detector generalizes under distribution shift.
We train the detector on each of the four datasets and evaluate it on the remaining datasets.
We also train on the union of all datasets (\textit{All}) and test on each dataset.
Unlike the main experiments, where we select the best layer per setting, here we fix layer 19 of Qwen-2.5-7B-Instruct for all experiments, since tuning the layer on the target dataset would violate the out-of-distribution setting.
\cref{fig:combined_ood} reports the AUROC for each train-test pair.

As the heatmaps show, the detector exhibits strong out-of-distribution generalization in both settings.
For question-only detection, most off-diagonal entries remain in the 72--88 AUROC range; for question-plus-answer, this improves to 75--93.
Notably, training on the union of all datasets (\textit{All}) yields strong and stable performance across all four targets, indicating that the learned features are dataset-agnostic rather than overfitting to source-specific patterns.
These results suggest that practitioners can train a single detector on available labeled data and deploy it across diverse query distributions without retraining.

\subsection{Accuracy after Hallucination Removed}

\begin{figure}[ht]
    \centering
    \includegraphics[width=0.9\columnwidth]{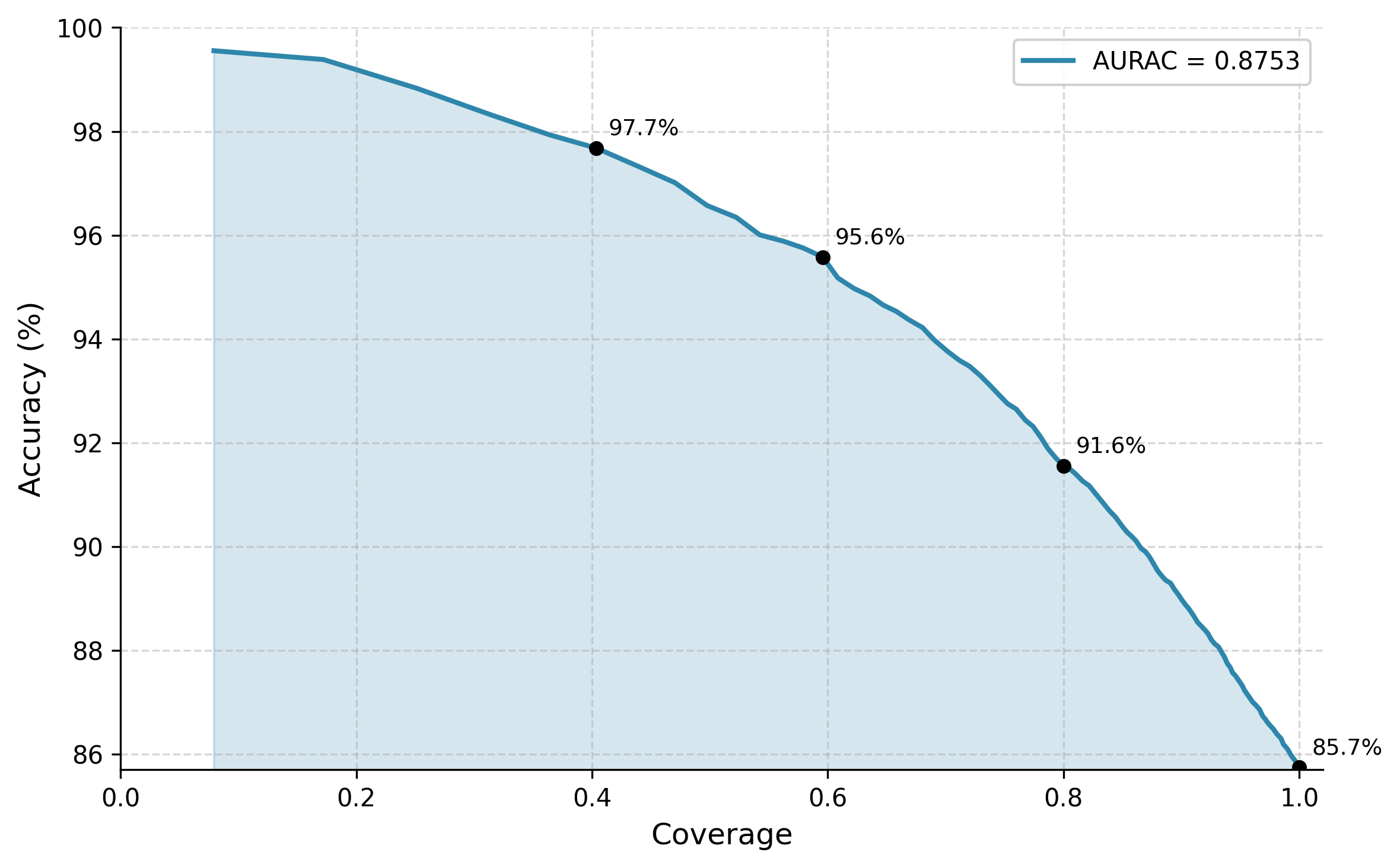}
    \caption{Rejection-Accuracy Curve (RAC) for Qwen-2.5-7B on TriviaQA using question token representations. Rejecting low-confidence samples substantially improves accuracy on retained samples.}
    \label{fig:rac}
\end{figure}

Beyond binary hallucination detection, our method can be used for \textit{selective prediction}: abstaining from answering when the model is likely to hallucinate.
In this experiment, we use only question token representations, allowing the detector to evaluate hallucination risk \textit{before} the model generates any answer, introducing zero additional latency.
We use the detector's output probability as a confidence score and reject samples below a threshold.
\cref{fig:rac} shows the rejection-accuracy curve (RAC) for Qwen-2.5-7B on TriviaQA, where we vary the confidence threshold and measure accuracy on the retained samples.

At full coverage, the model achieves 85.7\% accuracy.
By rejecting the 20\% least confident samples, accuracy rises to 91.6\%.
At 60\% coverage, accuracy reaches 95.6\%, and at 40\% coverage, it climbs to 97.7\%.
The area under the RAC (AURAC = 0.8753) quantifies this strong alignment between detector confidence and correctness.

This capability supports the LLM router architecture described in \cref{fig:pipeline}: queries flagged as high-risk can be routed to more reliable pipelines, while confident queries proceed without additional latency.
The steep accuracy gains at moderate rejection rates suggest that even conservative abstention thresholds can substantially improve system reliability.

\section{Ablation Studies}

To understand the design choices underlying our hallucination detector, we conduct several ablation studies.
In particular, we explore:
(1) which transformer layers provide the most informative representations for hallucination detection,
(2) how model scale influences classification and ranking quality,
(3) how different detector architectures and pooling strategies impact performance, and
(4) how detection performance varies with the number of answer tokens available at inference time.
Unless otherwise stated, all ablations are performed using Qwen2.5-7B-Instruct.

\subsection{Effect of Intermediate Layers on Performance}
\label{sec:layer-ablation}


\begin{figure}[ht]
    \centering
    \includegraphics[width=0.9\columnwidth]{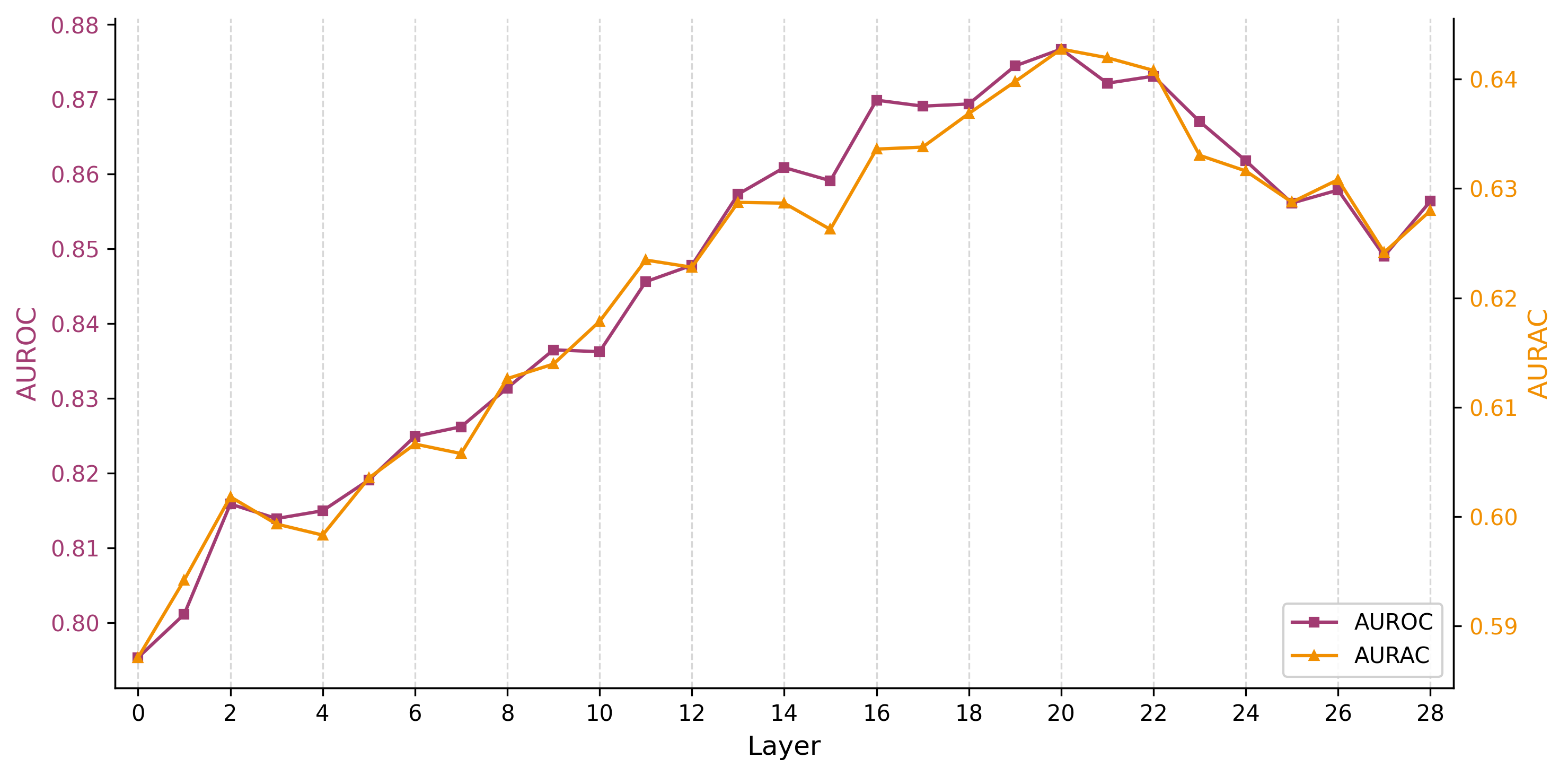}
    \caption{AUROC and AURAC across layers 0--28 for Qwen2.5-7B on WebQuestions. Both metrics peak around layer 20.}
    \label{fig:layer-ablation}
\end{figure}

As discussed in Section~\ref{sec:method}, we hypothesize that intermediate layers are more informative than the final layer for hallucination detection.
To validate this and identify the optimal layer, we train separate detectors on hidden states extracted from each layer and compare their performance.
This design is motivated by prior work showing that different transformer layers encode qualitatively different information~\citep{geometryoftruth, honestyalignment}.

\cref{fig:layer-ablation} shows AUROC and AURAC across layers 0--28 of Qwen2.5-7B.
Both metrics show an overall upward trend with minor fluctuations, peaking around layer 20 (AUROC 0.877, AURAC 0.643), before gradually declining in later layers.
This pattern suggests that intermediate-to-late layers provide the strongest signals for detecting hallucinations.
We attribute the decline in final layers to the projection toward token space, which discards internal confidence signals that are unnecessary for next-token prediction.
We observe similar trends for LLaMA-2-7B and Gemma-3-4B (see \cref{sec:ablation_study_2}).

\subsection{Effect of Model Scale}
\label{sec:model-scale}


\begin{figure}[ht]
    \centering
    \includegraphics[width=0.9\columnwidth]{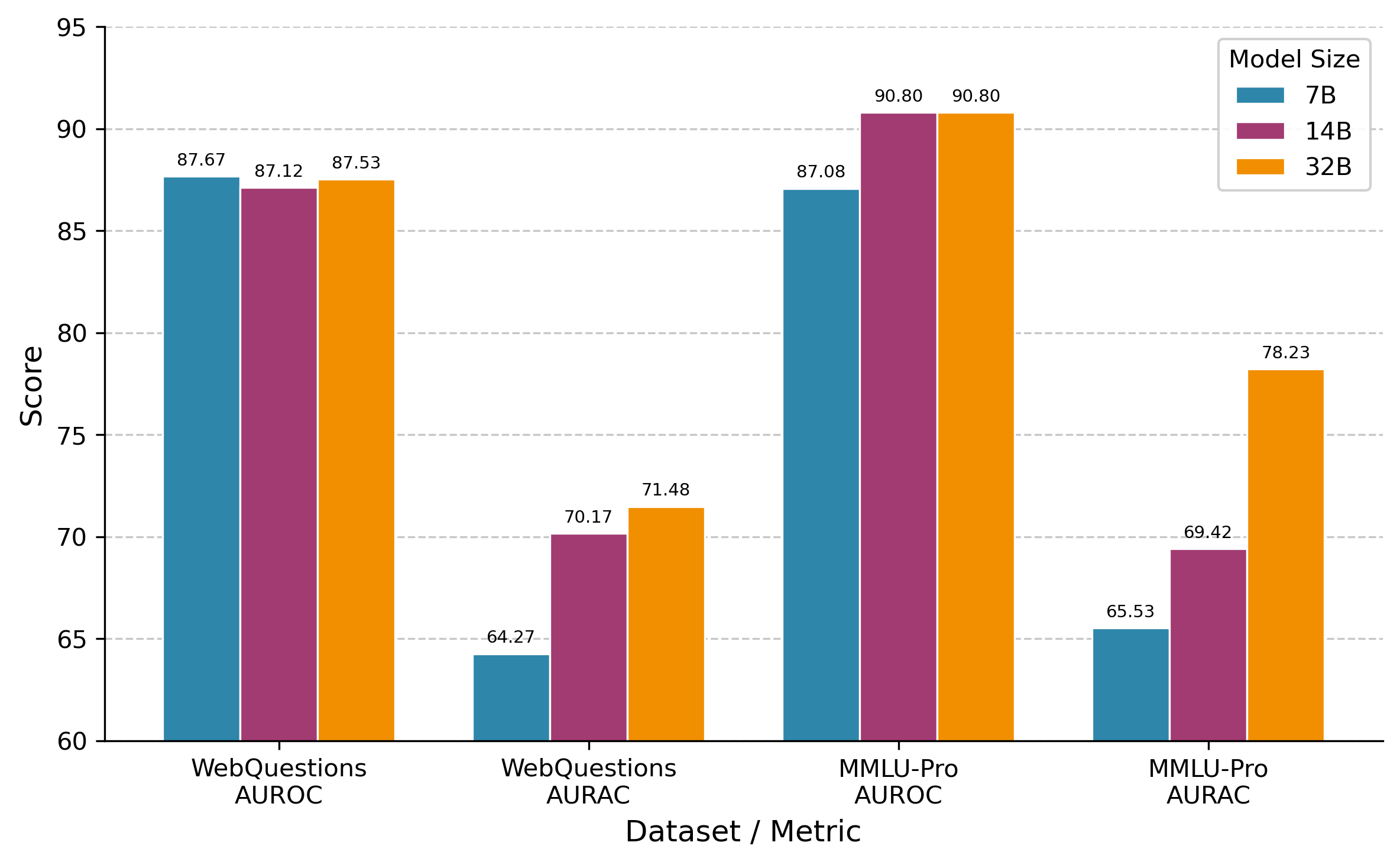}
    \caption{Effect of model scale (7B, 14B, 32B) on AUROC and AURAC. AUROC remains stable on WebQuestions and improves on MMLU-Pro. AURAC improves consistently on both datasets.}
    \label{fig:model-scale}
\end{figure}



In \cref{sec:intro}, we hypothesized that larger models exhibit a greater gap between internal representations and surface outputs, making our detector more effective.
To verify this, we evaluate Qwen2.5 models at 7B, 14B, and 32B on WebQuestions and MMLU-Pro.
As shown in \cref{fig:model-scale}, AUROC remains stable on WebQuestions and improves on MMLU-Pro.
AURAC improves consistently, with the 32B model achieving a 12.7-point gain over 7B on MMLU-Pro.
Both trends confirm that larger models yield stronger detection signals and better-calibrated confidence scores, supporting our hypothesis.
Given these scaling trends, we expect our method to be even more effective on frontier models with hundreds of billions of parameters.

\subsection{Effect of Detector Architecture}


\begin{figure}[ht]
    \centering
    \includegraphics[width=0.9\columnwidth]{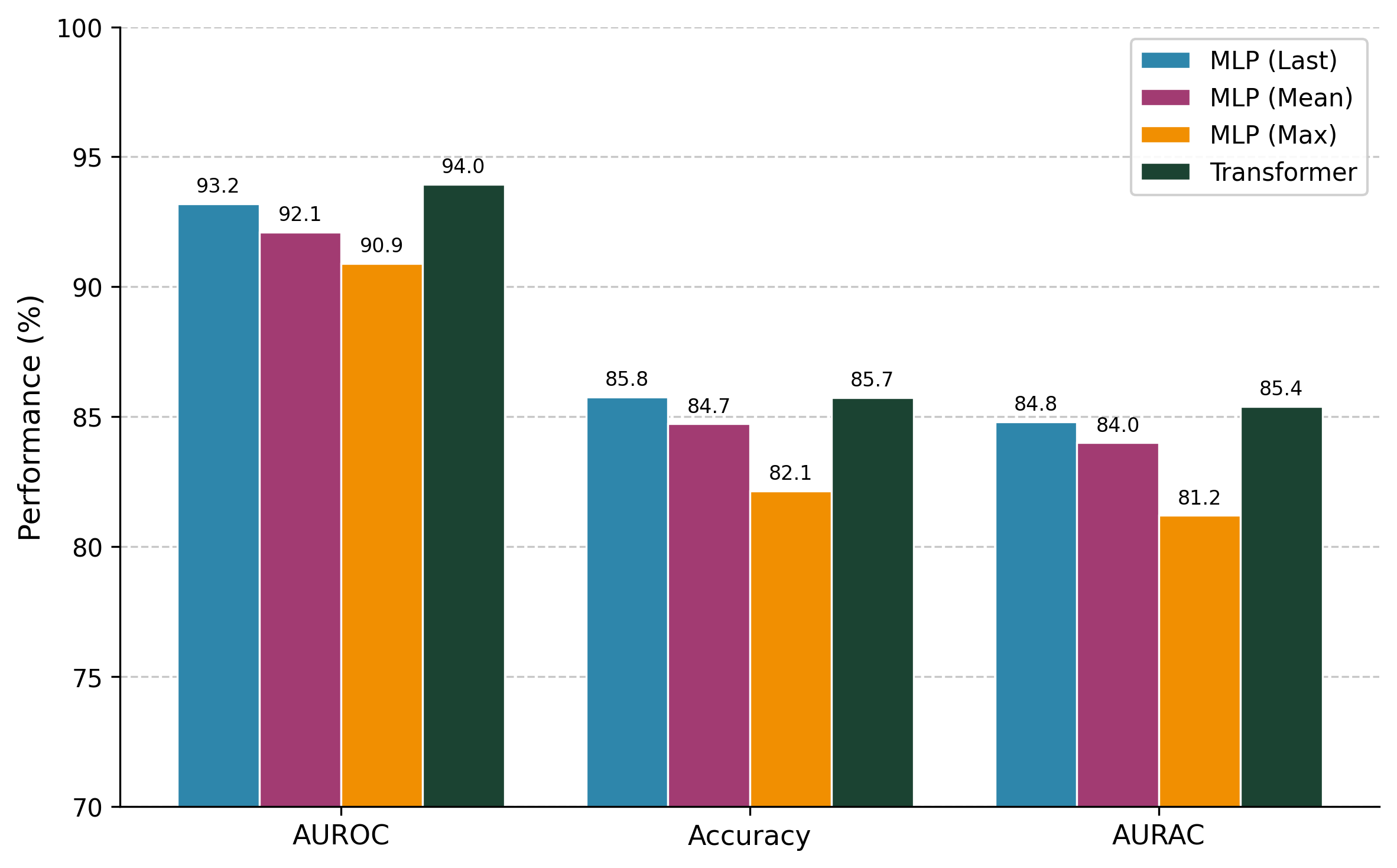}
    \caption{Comparison of detector architectures on TriviaQA (Qwen2.5-7B, layer 19). Last-token pooling performs best among MLP variants; the transformer architecture with attention pooling achieves the highest overall performance.}
    \label{fig:architecture-ablation}
\end{figure}

In \cref{sec:method}, we described two candidate architectures for the detector: an MLP with input-stage pooling and a lightweight transformer with attention pooling (\cref{tab:architecture}).
For the MLP, multiple pooling strategies are possible: last token, mean pooling, and max pooling.
Here we compare these design choices to identify the best configuration.
All experiments use hidden states from layer 19 of Qwen2.5-7B on TriviaQA.

As shown in \cref{fig:architecture-ablation}, among MLP variants, last-token pooling performs best, with the largest gain in Accuracy (3.6 points over max pooling) and improvements of 1.1--2.3 points in AUROC and 0.8--3.6 points in AURAC.
The transformer architecture further improves upon this, achieving comparable Accuracy while gaining 0.8 points in AUROC and 0.6 points in AURAC.
Based on these results, we adopt the transformer architecture in our main experiments.

\subsection{Effect of Answer Token Availability}

\begin{figure}[ht]
    \centering
    \includegraphics[width=0.9\columnwidth]{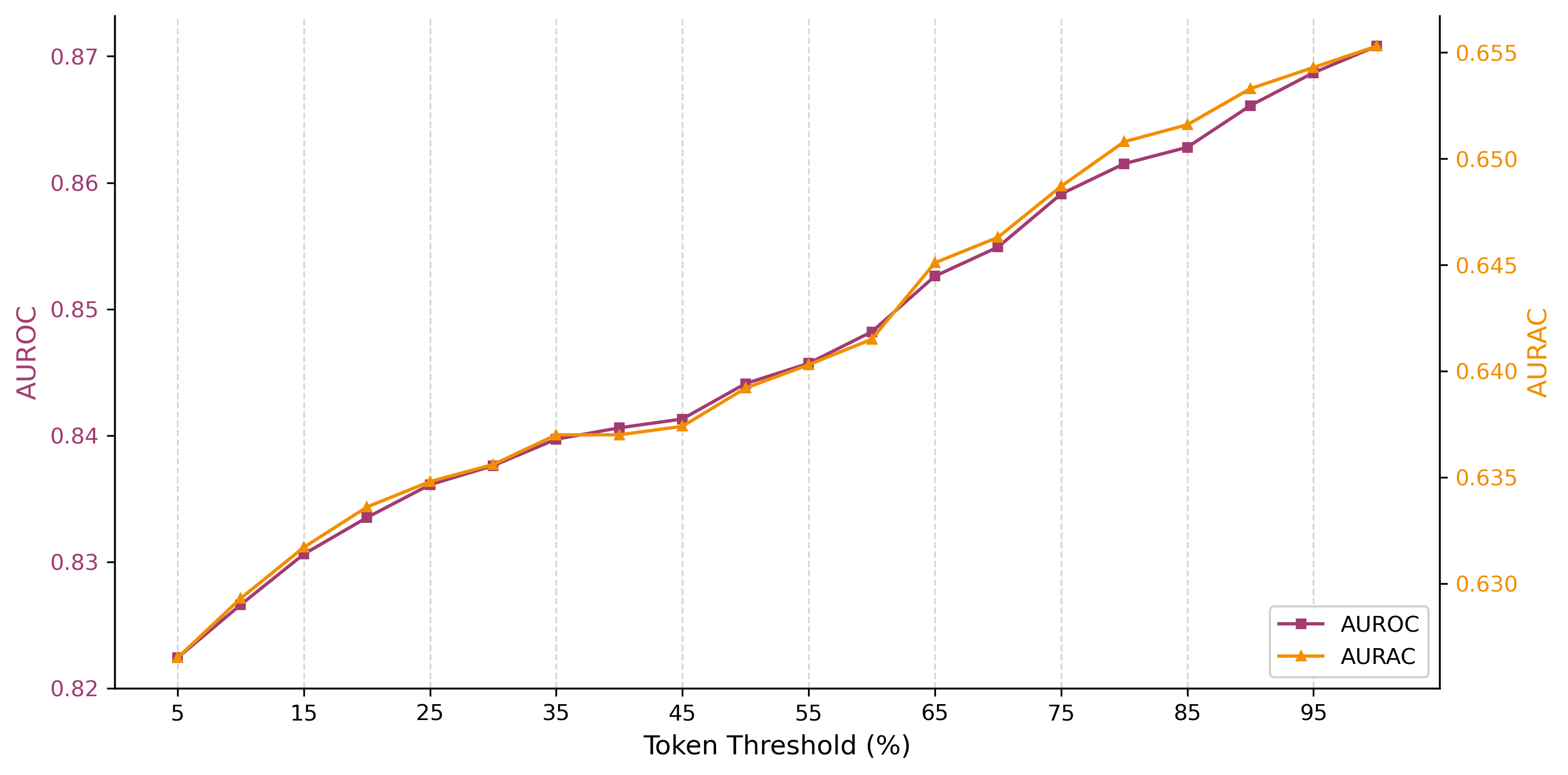}
    \caption{Effect of answer token availability on detection performance (Qwen2.5-7B, MMLU-Pro). Both AUROC and AURAC increase monotonically as more answer tokens are included.}
    \label{fig:tok_thresholds}
\end{figure}

Our main experiments compare question-only and question-plus-answer settings.
Here we explore an interesting setting: what if we evaluate the detector \textit{during} generation, using only the answer tokens produced so far?
We train a detector on question plus all answer tokens, then evaluate using only the first $x\%$, where $x$ ranges from 5\% to 100\%.
We use MMLU-Pro for this ablation because its answers involve multi-step reasoning, unlike TriviaQA and NQ-Open, where answers are often single tokens.

As shown in \cref{fig:tok_thresholds}, both AUROC and AURAC increase monotonically with the token threshold, exhibiting highly consistent behavior across the entire range.
AUROC improves by 4.9 points and AURAC by 2.8 points from 5\% to 100\% tokens.
This confirms that answer tokens carry cumulative signals for hallucination detection.
This enables flexible deployment.
Beyond question-only detection that runs fully in parallel with generation, practitioners can trigger detection partway through the response, achieving higher accuracy without incurring additional latency, which is critical for real-time user experience.

\section{Discussion}

\paragraph{Correctness as a Proxy for Hallucination}
We use correctness as a proxy for hallucination, which is standard practice in this field but warrants discussion.
Strictly speaking, our method detects cases where the model is uncertain yet still produces an output, rather than all types of hallucinations.
Models can produce errors for multiple reasons.
First, a lack of relevant knowledge, which our method can detect.
Second, systematic errors caused by incorrect training data, which our method cannot detect.
Third, logical errors during reasoning, which our method cannot detect.
Conflating these distinct mechanisms is not appropriate.
Furthermore, for ambiguous questions, underspecified queries, or questions with multiple valid answers, an incorrect label does not necessarily indicate hallucination.
Our datasets (TriviaQA, NQ-Open, MMLU-Pro, WebQuestions) primarily consist of factoid QA with well-defined answers, where this limitation has less impact.
Therefore, our reported results can be viewed as a lower bound on detection capability.
The fact that we achieve strong performance despite correctness, including error types that our method is not designed to detect, suggests that model uncertainty is a major source of LLM errors.

\paragraph{Data Imbalance}
Some model-dataset combinations exhibit severe label imbalance, as indicated by * in \cref{tab:mainresults}.
For example, LLaMA-2 achieves only 15.4\% accuracy on MMLU-Pro, meaning that correct answers constitute a small minority of the training data.
Although AUROC is more robust to class imbalance than accuracy (since it measures ranking quality rather than classification at a fixed threshold), results under extreme imbalance should still be interpreted with caution.
Notably, on settings with more balanced labels (such as Qwen across all datasets), our method achieves the strongest results across all benchmarks, suggesting that the method itself is effective and the weaker performance on imbalanced settings is attributable to data limitations rather than methodological flaws.

\paragraph{Limitations}
Our experiments focus on single-turn, closed-book question-answering tasks.
We do not test on multi-turn dialogues, tool use, code generation, or summarization.
We choose QA because it is a standard benchmark for hallucination detection research with clear ground-truth labels for evaluation.
Adapting our method to other scenarios requires collecting corresponding training data; in long-context settings, compression or sliding-window strategies may be needed to select informative representations.
The core assumption of our method (that intermediate layers encode uncertainty signals) is independent of task type, and we expect it to generalize to other generation tasks, though this requires further validation.

\paragraph{Scaling Trends}
In \cref{sec:model-scale} (\cref{fig:model-scale}), we find that our method performs better when applied to larger models.
A similar phenomenon has been reported in prior work~\cite{haloscope, languagemodelsmostlyknow}.
We therefore conjecture that our method generalizes to stronger, larger LLMs.
One possible explanation is that larger models have greater capacity and may perform richer latent computation beyond next-token prediction.
As a result, the final text can be less faithful to the intermediate state.
This can increase the gap between intermediate representations and surface outputs, making intermediate features more informative for our detector, thereby improving its effectiveness on larger models.

\paragraph{Future Directions}
Beyond hallucination detection and selective prediction, our method may benefit reinforcement learning for LLMs.
In tree search algorithms such as Monte Carlo tree search (MCTS), the uncertainty signal from our detector could guide exploration: branches where the model is uncertain warrant more exploration, while confident branches can be pruned early.
This could help prevent the model from converging to local minima driven by overconfident but incorrect predictions, leading to more stable policy optimization.

\section*{Impact Statement}

This paper presents work whose goal is to advance the field of Machine Learning. There are many potential societal consequences of our work, none of which we feel must be specifically highlighted here.

\section*{Acknowledgments}
HY was partially supported by the US National Science Foundation under awards IIS-2520978, GEO/RISE-5239902, the Office of Naval Research Award N00014-23-1-2007, DOE (ASCR) Award DE-SC0026052, and the DARPA D24AP00325-00.
Approved for public release; distribution is unlimited.

\bibliography{example_paper}

@misc{qwen3,
      title={Qwen3 Technical Report},
      author={An Yang and Anfeng Li and Baosong Yang and Beichen Zhang and Binyuan Hui and Bo Zheng and Bowen Yu and Chang Gao and Chengen Huang and Chenxu Lv and Chujie Zheng and Dayiheng Liu and Fan Zhou and Fei Huang and Feng Hu and Hao Ge and Haoran Wei and Huan Lin and Jialong Tang and Jian Yang and Jianhong Tu and Jianwei Zhang and Jianxin Yang and Jiaxi Yang and Jing Zhou and Jingren Zhou and Junyang Lin and Kai Dang and Keqin Bao and Kexin Yang and Le Yu and Lianghao Deng and Mei Li and Mingfeng Xue and Mingze Li and Pei Zhang and Peng Wang and Qin Zhu and Rui Men and Ruize Gao and Shixuan Liu and Shuang Luo and Tianhao Li and Tianyi Tang and Wenbiao Yin and Xingzhang Ren and Xinyu Wang and Xinyu Zhang and Xuancheng Ren and Yang Fan and Yang Su and Yichang Zhang and Yinger Zhang and Yu Wan and Yuqiong Liu and Zekun Wang and Zeyu Cui and Zhenru Zhang and Zhipeng Zhou and Zihan Qiu},
      year={2025},
      eprint={2505.09388},
      archivePrefix={arXiv},
      primaryClass={cs.CL},
      url={https://arxiv.org/abs/2505.09388},
}

@misc{gemini25,
      title={Gemini 2.5: Pushing the Frontier with Advanced Reasoning, Multimodality, Long Context, and Next Generation Agentic Capabilities},
      author={Gheorghe Comanici and Eric Bieber and Mike Schaekermann and Ice Pasupat and Noveen Sachdeva and Inderjit Dhillon and Marcel Blistein and Ori Ram and Dan Zhang and Evan Rosen and Luke Marris and Sam Petulla and others},
      year={2025},
      eprint={2507.06261},
      archivePrefix={arXiv},
      primaryClass={cs.CL},
      url={https://arxiv.org/abs/2507.06261},
}

@misc{anthropic2025biology,
  title        = {On the Biology of a Large Language Model},
  author       = {Lindsey, Jack and Gurnee, Wes and Ameisen, Emmanuel and Chen, Brian and Pearce, Adam and Turner, Nicholas L. and Citro, Craig and Abrahams, David and Carter, Shan and Hosmer, Basil and Marcus, Jonathan and Sklar, Michael and Templeton, Adly and Bricken, Trenton and McDougall, Callum and Cunningham, Hoagy and Henighan, Thomas and Jermyn, Adam and Jones, Andy and Persic, Andrew and Qi, Zhenyi and Thompson, T. Ben and Zimmerman, Sam and Rivoire, Kelley and Conerly, Thomas and Olah, Chris and Batson, Joshua},
  howpublished  = {\url{https://transformer-circuits.pub/2025/attribution-graphs/biology.html}},
  year         = {2025},
  note         = {Accessed: 2025-10-07; mechanistic interpretability study of Claude 3.5 Haiku}
}

@misc{anthropic2025cot,
      title={Reasoning Models Don't Always Say What They Think},
      author={Yanda Chen and Joe Benton and Ansh Radhakrishnan and Jonathan Uesato and Carson Denison and John Schulman and Arushi Somani and Peter Hase and Misha Wagner and Fabien Roger and Vlad Mikulik and Samuel R. Bowman and Jan Leike and Jared Kaplan and Ethan Perez},
      year={2025},
      eprint={2505.05410},
      archivePrefix={arXiv},
      primaryClass={cs.CL},
      url={https://arxiv.org/abs/2505.05410},
}

@misc{ren2023outofdistributiondetectionselectivegeneration,
      title={Out-of-Distribution Detection and Selective Generation for Conditional Language Models},
      author={Jie Ren and Jiaming Luo and Yao Zhao and Kundan Krishna and Mohammad Saleh and Balaji Lakshminarayanan and Peter J. Liu},
      year={2023},
      eprint={2209.15558},
      archivePrefix={arXiv},
      primaryClass={cs.CL},
      url={https://arxiv.org/abs/2209.15558},
}

@misc{Lexical-Similarity,
      title={Generating with Confidence: Uncertainty Quantification for Black-box Large Language Models},
      author={Zhen Lin and Shubhendu Trivedi and Jimeng Sun},
      year={2024},
      eprint={2305.19187},
      archivePrefix={arXiv},
      primaryClass={cs.CL},
      url={https://arxiv.org/abs/2305.19187},
}

@misc{EigenScore,
      title={INSIDE: LLMs' Internal States Retain the Power of Hallucination Detection},
      author={Chao Chen and Kai Liu and Ze Chen and Yi Gu and Yue Wu and Mingyuan Tao and Zhihang Fu and Jieping Ye},
      year={2024},
      eprint={2402.03744},
      archivePrefix={arXiv},
      primaryClass={cs.CL},
      url={https://arxiv.org/abs/2402.03744},
}

@misc{selfcheckGPT,
      title={SelfCheckGPT: Zero-Resource Black-Box Hallucination Detection for Generative Large Language Models},
      author={Potsawee Manakul and Adian Liusie and Mark J. F. Gales},
      year={2023},
      eprint={2303.08896},
      archivePrefix={arXiv},
      primaryClass={cs.CL},
      url={https://arxiv.org/abs/2303.08896},
}

@misc{semantic-entropy,
      title={Semantic Uncertainty: Linguistic Invariances for Uncertainty Estimation in Natural Language Generation},
      author={Lorenz Kuhn and Yarin Gal and Sebastian Farquhar},
      year={2023},
      eprint={2302.09664},
      archivePrefix={arXiv},
      primaryClass={cs.CL},
      url={https://arxiv.org/abs/2302.09664},
}

@article{semantic-entropy2,
  author    = {Farquhar, Sebastian and Kossen, Jannik and Kuhn, Lorenz and Gal, Yarin},
  title     = {Detecting hallucinations in large language models using semantic entropy},
  journal   = {Nature},
  year      = {2024},
  volume    = {630},
  number    = {8017},
  pages     = {625--630},
  doi       = {10.1038/s41586-024-07421-0},
  url       = {https://doi.org/10.1038/s41586-024-07421-0},
  issn      = {1476-4687}
}

@misc{steerllmlatentshallucination,
      title={Steer LLM Latents for Hallucination Detection},
      author={Seongheon Park and Xuefeng Du and Min-Hsuan Yeh and Haobo Wang and Yixuan Li},
      year={2025},
      eprint={2503.01917},
      archivePrefix={arXiv},
      primaryClass={cs.LG},
      url={https://arxiv.org/abs/2503.01917},
}

@misc{haloscope,
    title={HaloScope: Harnessing Unlabeled LLM Generations for Hallucination Detection},
    author={Xuefeng Du and Chaowei Xiao and Yixuan Li},
    year={2024},
    eprint={2409.17504},
    archivePrefix={arXiv},
    primaryClass={cs.LG}
}

@misc{harp,
    title={HARP: Hallucination Detection via Reasoning Subspace Projection},
    author={Junjie Hu and Gang Tu and ShengYu Cheng and Jinxin Li and Jinting Wang and Rui Chen and Zhilong Zhou and Dongbo Shan},
    year={2025},
    eprint={2509.11536},
    archivePrefix={arXiv},
    primaryClass={cs.CL}
}

@misc{languagemodelsmostlyknow,
      title={Language Models (Mostly) Know What They Know},
      author={Saurav Kadavath and Tom Conerly and Amanda Askell and Tom Henighan and Dawn Drain and Ethan Perez and Nicholas Schiefer and Zac Hatfield-Dodds and Nova DasSarma and Eli Tran-Johnson and Scott Johnston and Sheer El-Showk and Andy Jones and Nelson Elhage and Tristan Hume and Anna Chen and Yuntao Bai and Sam Bowman and Stanislav Fort and Deep Ganguli and Danny Hernandez and Josh Jacobson and Jackson Kernion and Shauna Kravec and Liane Lovitt and Kamal Ndousse and Catherine Olsson and Sam Ringer and Dario Amodei and Tom Brown and Jack Clark and Nicholas Joseph and Ben Mann and Sam McCandlish and Chris Olah and Jared Kaplan},
      year={2022},
      eprint={2207.05221},
      archivePrefix={arXiv},
      primaryClass={cs.CL},
      url={https://arxiv.org/abs/2207.05221},
}

@misc{truthuniversal,
      title={Truth is Universal: Robust Detection of Lies in LLMs},
      author={Lennart Bürger and Fred A. Hamprecht and Boaz Nadler},
      year={2024},
      eprint={2407.12831},
      archivePrefix={arXiv},
      primaryClass={cs.CL},
      url={https://arxiv.org/abs/2407.12831},
}

@misc{qwen2.5,
  title        = {Qwen2.5 Technical Report},
  author       = {An Yang and Baosong Yang and Beichen Zhang and Binyuan Hui and Bo Zheng and Bowen Yu and Chengyuan Li and Dayiheng Liu and Fei Huang and Haoran Wei and Huan Lin and Jian Yang and Jianhong Tu and Jianwei Zhang and Jianxin Yang and Jiaxi Yang and Jingren Zhou and Junyang Lin and Kai Dang and Keming Lu and Keqin Bao and Kexin Yang and Le Yu and Mei Li and Mingfeng Xue and Pei Zhang and Qin Zhu and Rui Men and Runji Lin and Tianhao Li and Tianyi Tang and Tingyu Xia and Xingzhang Ren and Xuancheng Ren and Yang Fan and Yang Su and Yichang Zhang and Yu Wan and Yuqiong Liu and Zeyu Cui and Zhenru Zhang and Zihan Qiu},
  year         = {2024},
  howpublished = {arXiv preprint arXiv:2412.15115v2},
  url          = {https://arxiv.org/abs/2412.15115v2}
}

@misc{llama2,
      title={LLaMA 2: Open Foundation and Fine-Tuned Chat Models},
      author={Hugo Touvron and Louis Martin and Kevin Stone and Peter Albert and Amjad Almahairi and Yasmine Babaei and Nikolay Bashlykov and Soumya Batra and Prajjwal Bhargava and Shruti Bhosale and Dan Bikel and Lukas Blecher and Cristian Canton Ferrer and Moya Chen and Guillem Cucurull and David Esiobu and Jude Fernandes and Jeremy Fu and Wenyin Fu and Brian Fuller and Cynthia Gao and Vedanuj Goswami and Naman Goyal and Anthony Hartshorn and Ufuk Kirnap and Igor Kivlichan and Marie-Anne Lachaux and Thibaut Lavril and Jenya Lee and Diana Liskovich and Yinghai Lu and Yuning Mao and Xavier Martinet and Todor Mihaylov and Pushkar Mishra and Igor Molybog and Yixin Nie and Andrew Poulton and Jeremy Reizenstein and Rashi Rungta and Kalyan Saladi and Alan Schelten and Ruan Silva and Eric Michael Smith and Ranjan Subramanian and Xiaoqing Ellen Tan and Binh Tang and Ross Taylor and Adina Williams and Jian Xiang Kuan and Punit Singh Koura and Shruti Bhosale and Deepak Narayanan and Anastasios Angelidis and Vishaal Shankar and Thomas Wolf and Aurelien Rodriguez and Serge Stojanov and Guillaume Lample and Tim Rocktäschel and Armand Joulin and Piotr Bojanowski and Edouard Grave and Alexis Conneau},
      year={2023},
      eprint={2307.09288},
      archivePrefix={arXiv},
      primaryClass={cs.CL},
      url={https://arxiv.org/abs/2307.09288}
}

@misc{gemma3,
      title={Gemma 3 Technical Report},
      author={Gemma Team},
      year={2025},
      eprint={2503.19786},
      archivePrefix={arXiv},
      primaryClass={cs.CL},
      url={https://arxiv.org/abs/2503.19786}
}

@misc{triviaqa,
      title={TriviaQA: A Large Scale Distantly Supervised Challenge Dataset for Reading Comprehension},
      author={Mandar Joshi and Eunsol Choi and Daniel S. Weld and Luke Zettlemoyer},
      year={2017},
      eprint={1705.03551},
      archivePrefix={arXiv},
      primaryClass={cs.CL},
      url={https://arxiv.org/abs/1705.03551},
}

@article{nqopen,
  title={Natural Questions: A Benchmark for Question Answering Research},
  author={Tom Kwiatkowski and Jennimaria Palomaki and Olivia Redfield and Michael Collins and Ankur P. Parikh and Chris Alberti and Danielle Epstein and Illia Polosukhin and Jacob Devlin and Kenton Lee and Kristina Toutanova and Llion Jones and Matthew Kelcey and Ming-Wei Chang and Andrew M. Dai and Jakob Uszkoreit and Quoc Le and Slav Petrov},
  journal={Transactions of the Association for Computational Linguistics},
  volume={7},
  pages={452--466},
  year={2019},
  url={https://aclanthology.org/Q19-1026},
}

@misc{mmlupro,
      title={MMLU-Pro: A More Robust and Challenging Multi-Task Language Understanding Benchmark},
      author={Yubo Wang and Xueguang Ma and Ge Zhang and Yuansheng Ni and Abhranil Chandra and Shiguang Guo and Weiming Ren and Aaran Arulraj and Xuan He and Ziyan Jiang and Tianle Li and Max Ku and Kai Wang and Alex Zhuang and Rongqi Fan and Xiang Yue and Wenhu Chen},
      year={2024},
      eprint={2406.01574},
      archivePrefix={arXiv},
      primaryClass={cs.CL},
      url={https://arxiv.org/abs/2406.01574}
}

@inproceedings{webquestions,
    title={Semantic Parsing on Freebase from Question-Answer Pairs},
    author={Jonathan Berant and Andrew Chou and Roy Frostig and Percy Liang},
    booktitle={Proceedings of the 2013 Conference on Empirical Methods in Natural Language Processing},
    pages={1533--1544},
    year={2013},
    url={https://aclanthology.org/D13-1160}
}

@misc{geometryoftruth,
      title={The Geometry of Truth: Layer-wise Semantic Dynamics for Hallucination Detection in Large Language Models},
      author={Amir Hameed Mir},
      year={2025},
      eprint={2510.04933},
      archivePrefix={arXiv},
      primaryClass={cs.CL},
      url={https://arxiv.org/abs/2510.04933},
}

@misc{honestyalignment,
      title={Annotation-Efficient Universal Honesty Alignment},
      author={Shiyu Ni and Keping Bi and Jiafeng Guo and Minghao Tang and Jingtong Wu and Zengxin Han and Xueqi Cheng},
      year={2025},
      eprint={2510.17509},
      archivePrefix={arXiv},
      primaryClass={cs.CL},
      url={https://arxiv.org/abs/2510.17509},
}
\bibliographystyle{icml2026}

\newpage
\appendix
\onecolumn

\section{Correctness Evaluation Prompt}
\label{sec:prompt}

We use the following prompt template to obtain correctness labels from GPT-4o.

\begin{lstlisting}
Evaluate whether the generated answer is CORRECT or INCORRECT.
Question: {question}
Ground truth: {answer}
Generated: {generated_text}
A generated answer is CORRECT if it expresses the same meaning
as the ground truth, without introducing incorrect, conflicting,
or extra information. Otherwise, it is INCORRECT.
Respond with EXACTLY "true" or "false".
\end{lstlisting}

\section{Dataset Statistics}
\label{sec:dataset-statistics}

\begin{table}[ht]
    \centering
    \caption{Correctness label statistics across models and benchmarks. Accuracy (\%) is shown for each model-dataset pair.}
    \begin{tabular}{ll|cccc}
    \toprule
        Model & Metric & TriviaQA & NQ-Open & MMLU-Pro & WebQuestions \\
    \midrule
        \multirow{2}{*}{LLaMA-2-7B} & Count & 18,585 / 12,728 & 7,946 / 15,664 & 1,851 / 10,180 & 2,215 / 3,570 \\
        & Accuracy & 59.4\% & 33.7\% & 15.4\% & 38.3\% \\
    \midrule
        \multirow{2}{*}{Qwen2.5-7B} & Count & 17,743 / 13,570 & 7,287 / 16,323 & 4,403 / 7,629 & 2,153 / 3,645 \\
        & Accuracy & 56.7\% & 30.9\% & 36.6\% & 37.1\% \\
    \midrule
        \multirow{2}{*}{Gemma-3-4B} & Count & 14,945 / 16,368 & 5,537 / 18,037 & 3,264 / 8,767 & 1,835 / 3,975 \\
        & Accuracy & 47.7\% & 23.5 \% & 27.1\% & 31.6\% \\
    \bottomrule
    \end{tabular}
    \label{tab:dataset-statistics}
\end{table}

\section{Layer Ablation on Additional Models}
\label{sec:ablation_study_2}

\begin{figure}[ht]
    \centering
    \begin{subfigure}[b]{0.45\linewidth}
        \centering
        \includegraphics[width=\linewidth]{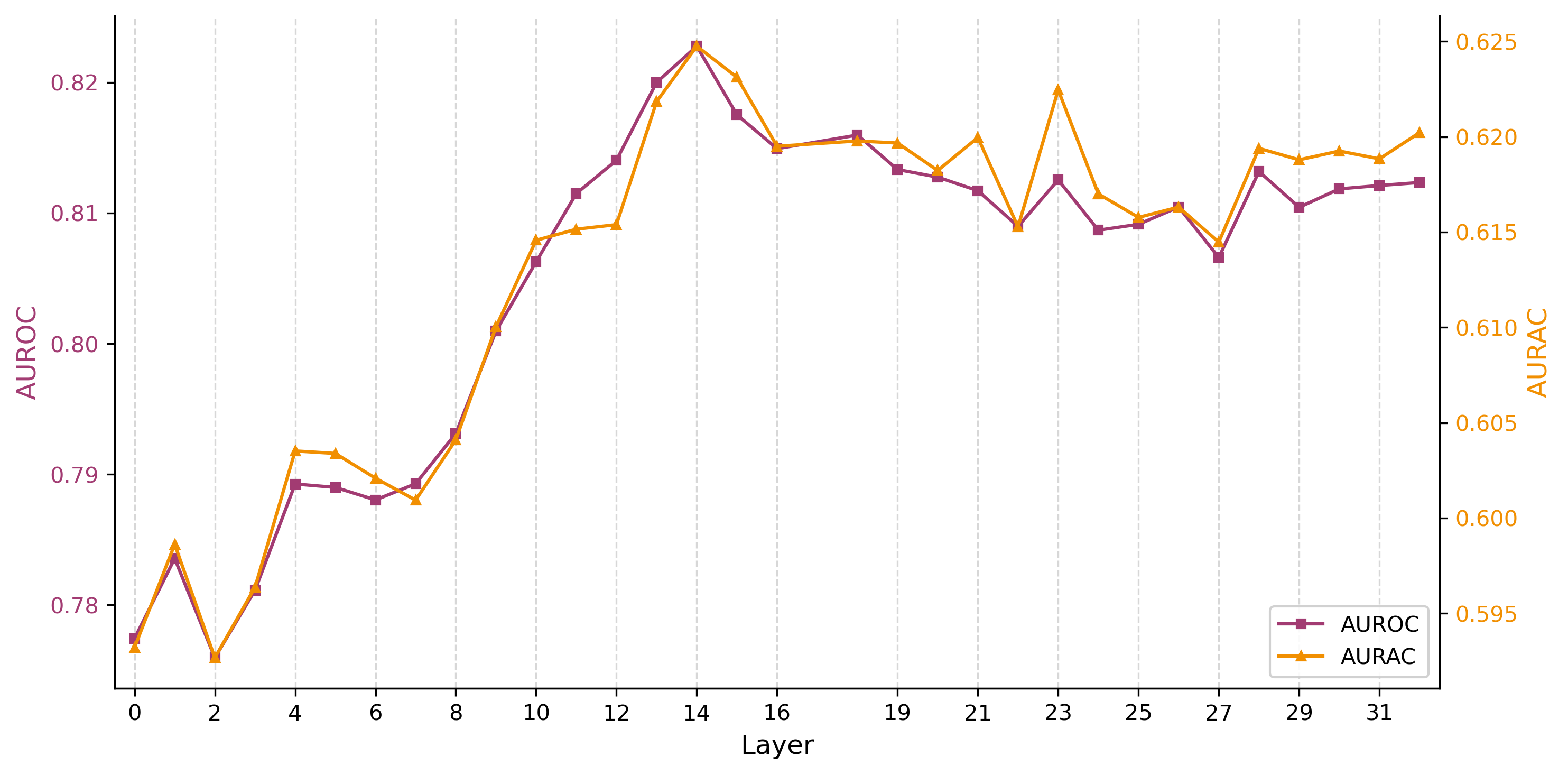}
        \caption{LLaMA-2-7B}
        \label{fig:ablation_llama}
    \end{subfigure}
    \hfill 
    \begin{subfigure}[b]{0.45\linewidth}
        \centering
        \includegraphics[width=\linewidth]{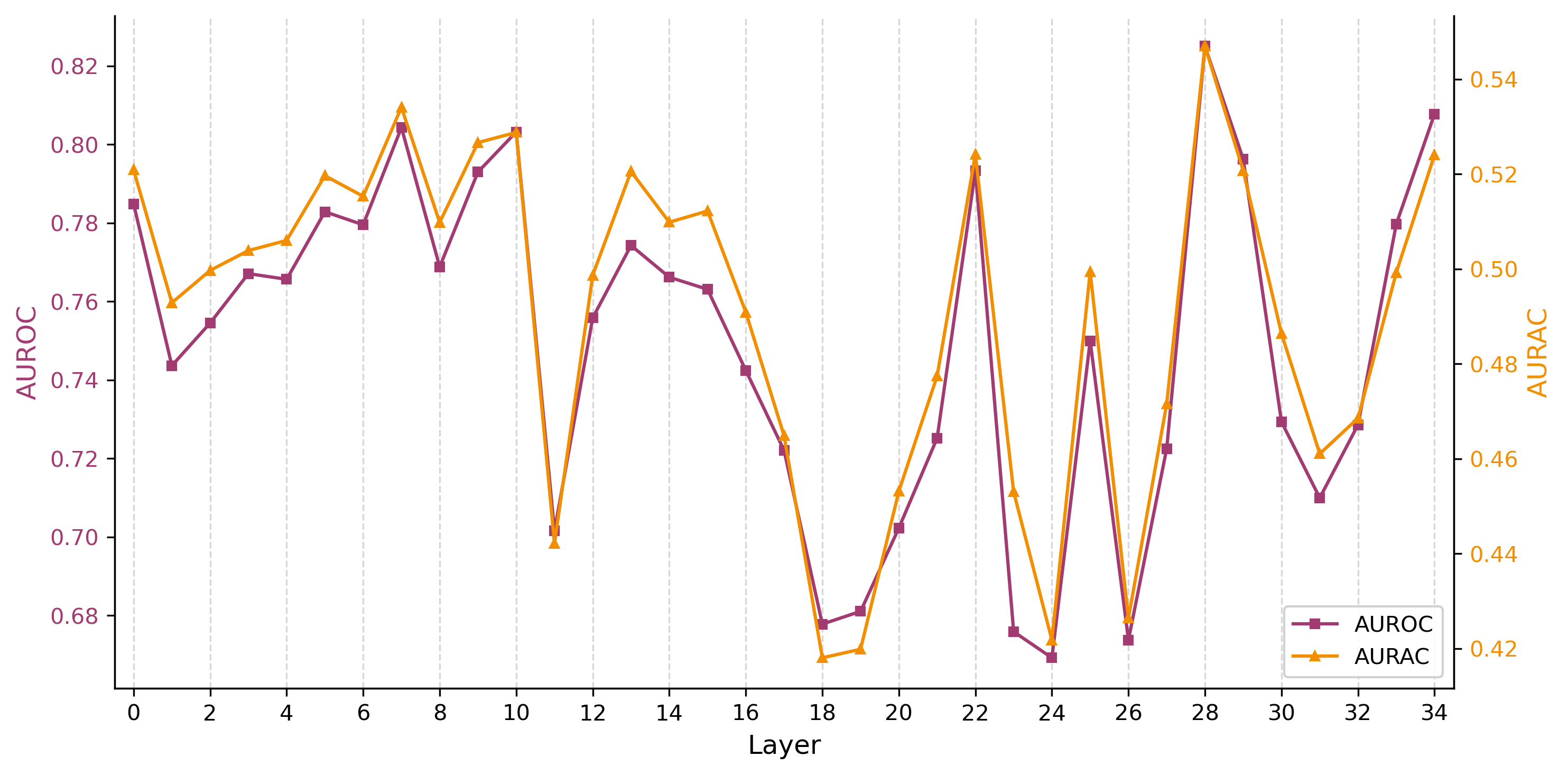}
        \caption{Gemma-3-4B}
        \label{fig:ablation_gemma}
    \end{subfigure}

    \caption{Layer ablation results for LLaMA-2-7B and Gemma-3-4B on WebQuestions. Peak performance occurs at intermediate layers.}
    \label{fig:layer_ablation_comparison}
\end{figure}

We extend the layer ablation study from \cref{sec:layer-ablation} to LLaMA-2-7B and Gemma-3-4B to verify that the trend observed in Qwen2.5-7B generalizes across model families.
\cref{fig:layer_ablation_comparison} shows AUROC and AURAC across all layers for both models on the WebQuestions dataset.

For LLaMA-2-7B (32 layers), performance peaks around layer 14, with AUROC and AURAC declining in both earlier and later layers.
For Gemma-3-4B (35 layers), the peak occurs around layer 28.
Interestingly, the Gemma-3-4B curve exhibits notable fluctuations across layers, suggesting that different layers may encode qualitatively different information; this warrants further investigation.
In both cases, intermediate-to-late layers outperform early layers and the final layers, confirming that the optimal layer for hallucination detection is not the last layer but rather an intermediate one.
This pattern aligns with findings in interpretability literature suggesting that middle layers encode richer semantic representations.

\end{document}